\documentclass[10pt,twocolumn,letterpaper]{article}

\usepackage{cvpr}              

\definecolor{cvprblue}{rgb}{0.21,0.49,0.74}
\usepackage[pagebackref,breaklinks,colorlinks,allcolors=cvprblue]{hyperref}
\usepackage{algorithm}
\usepackage[noend]{algpseudocode}
\usepackage{xcolor}

\def\paperID{3} 
\def\confName{CVPR}
\def\confYear{2025}

\title{Bounding Box-Guided Diffusion for Synthesizing \\ Industrial Images and Segmentation Maps}

\author{
Emanuele Caruso\thanks{Equal contribution.}~, Alessandro Simoni\footnotemark[1]~, Francesco Pelosin\footnotemark[1]\\
Covision Lab\\
Brixen, South Tyrol, Italy\\
{\tt\small name.surname@covisionlab.com}
}

\begin{document}
\maketitle
\begin{abstract}
Synthetic dataset generation in Computer Vision, particularly for industrial applications, is still underexplored. Industrial defect segmentation, for instance, requires highly accurate labels, yet acquiring such data is costly and time-consuming. To address this challenge, we propose a novel diffusion-based pipeline for generating high-fidelity industrial datasets with minimal supervision. Our approach conditions the diffusion model on enriched bounding box representations to produce precise segmentation masks, ensuring realistic and accurately localized defect synthesis. Compared to existing layout-conditioned generative methods, our approach improves defect consistency and spatial accuracy. We introduce two quantitative metrics to evaluate the effectiveness of our method and assess its impact on a downstream segmentation task trained on real and synthetic data. Our results demonstrate that diffusion-based synthesis can bridge the gap between artificial and real-world industrial data, fostering more reliable and cost-efficient segmentation models. The code is publicly available at \url{https://github.com/covisionlab/diffusion_labeling}.
\end{abstract}    
\section{Introduction}
\label{sec:intro}  
Dataset synthesis has gained significant importance in recent years, particularly within the Natural Language Processing (NLP) community, where we witnessed major improvements in both academic and industrial applications \cite{tencent_synth, self_instruct, self_play_ft}. These methods have proven especially valuable in scenarios where collecting and annotating real-world data is expensive or impractical.

\begin{figure*}
    \centering
    \includegraphics[width=\linewidth]{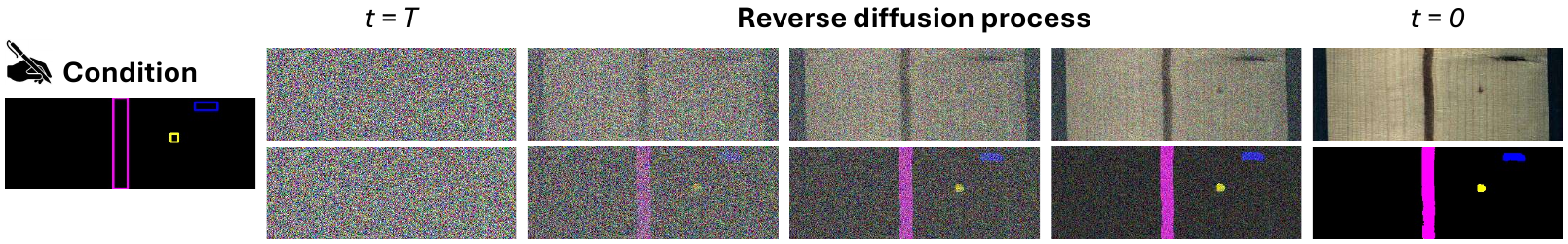}
    \caption{Overview of the proposed diffusion-based approach that generates both RGB and segmentation map in industrial setting.}
    \label{fig:overview}
\end{figure*}

In contrast, dataset synthesis in Computer Vision remains an emerging field and its usage is still under study \cite{quality_gen, real_vs_synth}. Its potential to reduce labeling costs and mitigate data scarcity constitute an appealing property for the deep learning paradigm. Despite its potential, the field remains relatively underexplored compared to its NLP counterpart. This is particularly true in domains where acquiring precise labeled data is both costly and time-consuming, such as industrial inspection, medical imaging, and remote sensing. In these domains, even small inaccuracies in annotation can significantly impact model performance, making synthetic data generation a compelling alternative.

Most of the recent research in synthetic data for vision has focused on text-to-image generation \cite{li2024genai, zhang2024object, sun2023dreamsync}, leveraging generative models to create realistic visuals from textual descriptions. While these advancements have paved the way for creative applications and content generation, their direct applicability to real-world industrial settings remains limited. Industrial datasets, in particular, suffer from challenges such as class imbalances, labeling inconsistencies and high quality standards. These issues necessitate the development of tailored synthesis techniques capable of generating high-fidelity data hopefully with minimal manual intervention.

A critical challenge is the automatic creation of industrial dataset samples, where balancing efficiency with accuracy is difficult. Fully automated synthesis risks generating unrealistic or irrelevant samples, reducing the utility of the data. On the other hand, manual supervision, while improving accuracy, is often infeasible due to time and cost constraints — especially when dealing with complex imaging systems that go beyond human perception such as infrared imaging \cite{infrared_ships}. Industrial defect segmentation exemplifies this challenge, as it demands highly precise annotations to train reliable models.

To address these limitations, we propose a novel pipeline for generating realistic synthetic samples with cheap supervision. Our approach leverages diffusion models conditioned on human-provided bounding boxes to produce precise segmentation masks. 
By doing so, we unlock the generation of high-quality industrial datasets while exploiting human domain expertise but with a significant reduction in the burden of manual annotation.

In industrial settings, diffusion models have been employed for data synthesis only in classification tasks \cite{10317774}. However, to the best of our knowledge and according to a recent review \cite{review}, no existing work has addressed the challenge of synthetic dataset generation for semantic segmentation, generating high quality labels from inexpensive annotation.

We present a diffusion-based approach, depicted in Figure~\ref{fig:overview}, that generates RGB images and semantic maps leveraging an enriched bounding box representation as conditioning. We compare it with a modified state-of-the-art approach on layout-conditioned generation \cite{layoutdiffusion}. Our baseline exhibits superior consistency in generating defects within the provided bounding box annotations, making it preferable over existing generative pipelines. In this regard, we propose two metrics to quantitatively evaluate the obtained results. Ultimately, we provide some experiments showing the quality of the generated data by monitoring the performance of a downstream segmentation task trained on both real and synthetic data. Thus, we shed light on the potential of diffusion-based synthesis in bridging the gap between artificial and real-world industrial data, fostering more accurate and efficient computer vision models for segmentation.

To sum up, our main contributions are as follows:
\begin{itemize}
    \item We introduce a novel synthetic data generation pipeline that leverages diffusion models conditioned on human-provided bounding boxes to generate high-fidelity industrial dataset samples.
    \item The proposed approach, thanks to an enriched bounding box representation, ensures that the generated defects remain both realistic and accurately localized within the bounding box boundary, enhancing segmentation consistency.
    \item By reducing the reliance on manual labeling, our method significantly lowers the cost and time required for curating industrial datasets while maintaining high annotation quality.
    \item We propose two metrics and evaluate our approach against a state-of-the-art conditioned diffusion pipeline, demonstrating competitive performance and improved control over defect placement.
    \item Our findings highlight the potential of diffusion-based dataset synthesis to improve industrial defect segmentation models, unlocking the development of more robust computer vision solutions in real-world settings.
\end{itemize}

\section{Related Works}
\label{sec:related}

Synthetic data generation has been explored through various methodologies, each catering to specific domains and applications. 

\textbf{3D Game Engines.} One prevalent approach leverages 3D game engines such as Unreal Engine~\cite{unreal}, where meticulously crafted scenes or objects serve as high-fidelity proxies of reality. This method has been widely adopted, leading to the creation of extensive datasets and comprehensive frameworks \cite{synthia, VKITTI, infinigen2023infinite}, which have subsequently facilitated advancements in novel methodologies \cite{da_survey, sun20233d}.

\textbf{GAN / Diffusion.} Another powerful paradigm involves neural generative models. Techniques such as GANs \cite{gan} and diffusion models \cite{diffusion} have demonstrated remarkable efficacy in producing high-fidelity synthetic data. These models have found widespread applications, ranging from medical imaging \cite{guo2024maisi, medical_gan, FernandezPBTGVC22}, self-driving car research \cite{DBLP:conf/nips/0001VTN23, DBLP:conf/nips/PronovostGHWMWR23}, privacy preservation \cite{DBLP:conf/cvpr/KlempRWQL23} and finally in robotics, where has been investigated for pose estimation, as discussed in \cite{dr_gibboni_paper}.

\textbf{Foundation Models.} Recently, foundation models have also been explored for synthetic data generation. Notably, COSMOs \cite{cosmos} facilitates the creation of entire synthetic video sequences, while large vision-text models have been widely utilized for generative applications \cite{li2024genai, zhang2024object, sun2023dreamsync}.

\textbf{Conditioned Generation.} Our pipeline not only generates RGB images but also their correspondent labels. A related study \cite{satsynth}, proposes a method for end-to-end RGB and label generation for satellite data. While their approach is purely generative, ours allows human intervention, granting users the flexibility to place annotations as needed. This distinction enhances the control and accuracy of label generation.

Additionally, we consider \cite{layoutdiffusion}, a generative approach that conditions data synthesis on bounding boxes. We will compare our method with this approach in later sections to provide a comprehensive evaluation of our proposed framework.

\begin{figure*}
    \centering
    \includegraphics[width=\linewidth]{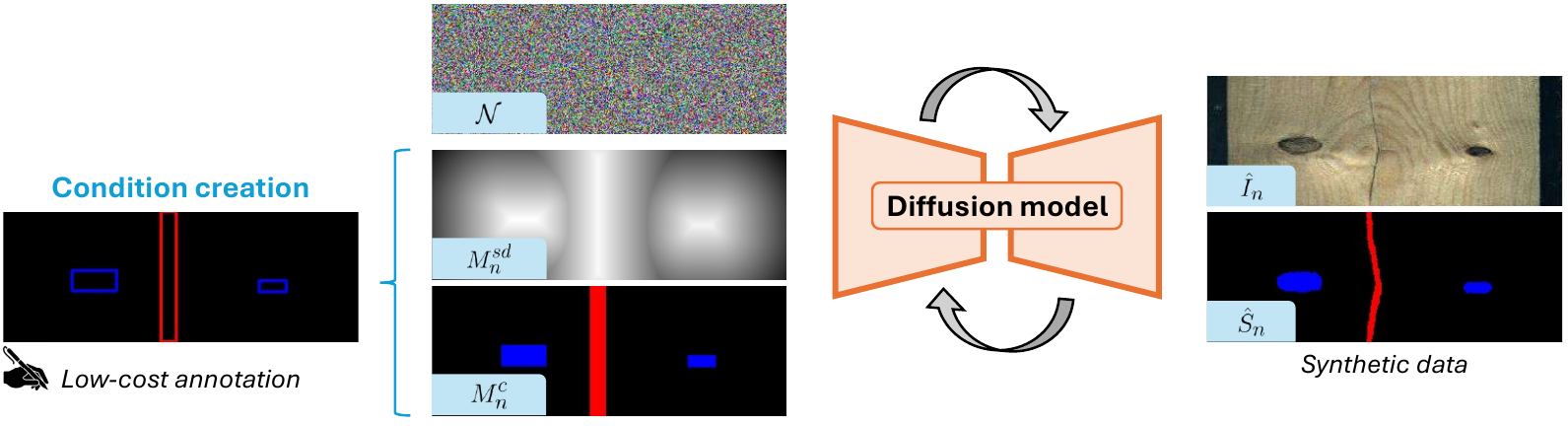}
    \caption{An overview of the proposed method: the user produces low-cost bounding box annotations which are then converted in two representations (BASD and C-BASD). Later, these encodings, are fed into the diffusion to condition the generation of both high quality RGB and segmentation masks of wood defects.}
    \label{fig:model}
\end{figure*}

\section{Method}
\label{sec:method}

In this section, we describe the proposed method shown in Figure~\ref{fig:model}.

\subsection{Problem statement}
In this work, we address the challenging task of semantic segmentation in an industrial setting. Since the lack of annotated data is very common, a way to tackle this problem is to augment the annotations with synthetic samples. Thus, we aim to adapt a conditional diffusion-based pipeline to denoise both an RGB image and its segmentation map as annotation.

Formally, we define a dataset $\mathcal{D}=\{(I_n, S_n, B_n) \mid n = 1, \dots, N\}$ where:
\begin{itemize}
    \item $I_n^{H \times W \times 3}$ is an RGB image,
    \item $S_n^{H \times W}$ is the corresponding segmentation map composed of the discrete pixels values $c_{ij} \in \{1,2,3,\dots,C\}$ where $C$ is the total number of classes,
    \item $B_n = \{b_k : (c, i_{min}, j_{min}, i_{max}, j_{max}), k=1,\dots,K\}$ is a tuple that identifies the class of the object and its bounding box location as the top left $(i_{min}, j_{min})$ and bottom right $(i_{max}, j_{max})$ corners.
\end{itemize}
Our method applies the diffusion process to the couple $(I_n, S_n)$ conditioned on $B_n$. In the following, we thoroughly describe how we preprocess the inputs and the training pipeline of the proposed method.

\subsection{Data preprocessing}
The first step is to process the segmentation map $S_n$ and the bounding boxes $B_n$ to allow the diffusion process to work with continuous values.

\textbf{Segmentation map.} Since the goal is to generate synthetic samples according to the joint probability $p(I_n, S_n)$, we need to make sure that these data are in the same continuous space $\mathbb{R}$. Drawing inspiration from \cite{analogbits,satsynth}, we convert the segmentation map into an analog bit representation. Formally, the pixelwise discrete segmentation values $c_{ij}$ are mapped into a binary code defined as 
\begin{equation}
    \textit{bin}: \{1, 2, 3 \dots, C\} \rightarrow \{0,1\}^{\lceil \log_2 C \rceil}
\end{equation}
After this encoding, the segmentation map dimension is $H \times W \times \lceil \log_2 C \rceil$. As proven by previous works \cite{analogbits}, this representation is more effective than one-hot encoding which is also less efficient in terms of number of channels in the presence of a high number of classes $C$. After the binary encoding, a normalization is applied to change the range from $[0,1]$ to $[-1, 1]$ which is the same of the RGB image $I_n$.

\textbf{Bounding box.} To condition the generation of the synthetic couple $(\hat{I}_n, \hat{S}_n)$ on the bounding boxes, we create an enriched representation of $B_n$ that encodes both spatial and class information. The spatial information is captured in terms of pixelwise encoding. Thus, we compute a Bounding Box-Aware Signed Distance (BASD) map $M_n^{d}$ that assigns to each pixel $(i,j)$ the minimum distance to the nearest bounding box boundary point. The distance value is positive inside a bounding box and negative outside. Moreover, a Bounding-Box Class (C-BASD) map $M_n^{c}$ is computed accordingly assigning to each positive value the corresponding class of the boundary point. We formally define the computation of $M_n^{d}$ and $M_n^{c}$ in Algorithm~\ref{alg:sdf-cm} and a visualization of the resulting maps can be seen in Figure~\ref{fig:model}.

\begin{algorithm}
    \caption{$M_n^{d}$ and $M_n^{c}$ computation. Comments in \textcolor{blue}{blue}.}\label{alg:sdf-cm}
    \begin{algorithmic}[1]
    \Require Bounding boxes $B_n$
    \Ensure $M_n^{d}$ of size $(H, W)$, $M_n^{c}$ of size $(H, W)$
    
    \State Initialize $M_n^{d} \gets +\infty$ for all pixels $p_{ij}$
    \State Initialize $M_n^{c} \gets 0$ for all pixels $p_{ij}$

    \For{each $b_k \in B_n$ with class $c$}
        \State \textcolor{blue}{Compute boundary pixels of $b_k$:}
        \State $\beta \gets \text{Boundary}(b_k)$    
        \For{each pixel $p_{ij}$}
            \State \textcolor{blue}{Compute distance to the closest boundary point:}
            \State $d_{\beta} \gets \min\limits_{(i_\beta, j_\beta) \in \beta} \sqrt{(i - i_\beta)^2 + (j - j_\beta)^2}$
            \State $d_{\beta} \gets d_{\beta} * \text{InOutSign}(p_{ij},b_k)$

            \State \textcolor{blue}{Update $M_n^{d}$ and $M_n^{c}$:}
            \If{$|d_{\beta_n}| < |M_n^{d}(p_{ij})|$}
                \State $M_n^{d}(p_{ij}) \gets d_{\beta}$
                \State $M_n^{c}(p_{ij}) \gets c$
            \EndIf
        \EndFor
    \EndFor    
    \end{algorithmic}
\end{algorithm}

Before concatenating these two representation maps to the couple $(I_n, S_n)$, the class map $M_n^{c}$ is encoded with the previously introduced analog bit paradigm obtaining an output dimension of $H \times W \times \lceil \log_2 C \rceil$. Our encoding assigns a single class per pixel but still handles overlapping bounding boxes. When two boxes overlap, the class map forms a structured pattern reflecting the overlap location instead of arbitrarily selecting one class. This allows the network to learn spatial relationships without needing explicit multi-label assignments, which a pure analog bit encoding can not achieve.

\begin{figure*}
    \centering
    \includegraphics[width=0.98\linewidth]{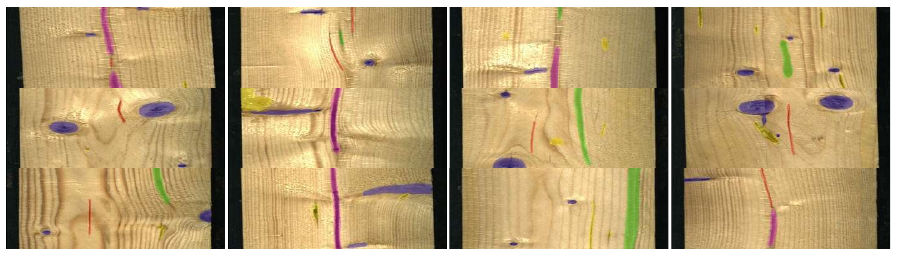}
    \caption{Some samples of the Wood Defect Detection \cite{openwood} with the semantic segmentation labels. The wood defects are the following: knot (blue), crack (red), quartzite (green), resin (yellow), marrow (magenta).}
    \label{fig:dataset}
\end{figure*}

\subsection{Conditioned Diffusion Model}
To synthesize realistic and structurally consistent images, we condition the denoising diffusion process on our enriched bounding box representation. A UNet architecture takes as input $(x_0, \, (M_n^{d}, M_n^{c}))$ where $x_0 = (I_n, S_n)$.
The output is the couple $(\hat{I}_n, \hat{S}_n)$ comprising of an RGB image plus its segmentation map with dimension $H \times W \times 3 + \lceil \log_2 C \rceil$.

Given a clean sample $x_0$, the forward diffusion process gradually adds Gaussian noise:
\begin{equation}
q(x_t \mid x_0) = \mathcal{N}(x_t; \sqrt{\alpha_t}\, x_0,\, (1 - \alpha_t) I),
\end{equation}
where $\alpha_t$ is the noise scheduling coefficient.
The reverse process learns to reconstruct $x_0$ while incorporating the structural constraints from the conditioning $(M_n^{d}, M_n^{c})$:
\begin{equation}
    \begin{aligned}
        &p_\theta(x_{t-1} \mid x_t, M_n^{d}, M_n^{c}) = \\
        &\mathcal{N} \big( x_{t-1}; \mu_\theta(x_t, t, M_n^{d}, M_n^{c}), \sigma_t^2 I \big).
    \end{aligned}
\end{equation}
where $\mu_\theta(x_t, t, M_n^{d}, M_n^{c})$ is the predicted denoised estimate and $\sigma_t$ is the variance of the noise distribution.

The diffusion model is trained by minimizing the noise prediction loss:
\begin{equation}
\mathbb{E}_{x_0, M_n^{d}, M_n^{c}, t, \epsilon} \left[ \|\epsilon - \epsilon_\theta(x_t, t, M_n^{d}, M_n^{c})\|^2 \right],
\end{equation}
with $\epsilon \sim \mathcal{N}(0, I)$ representing the injected Gaussian noise. This formulation ensures that the generated samples adhere to both the semantic structure encoded in the segmentation and the spatial constraints provided as bounding box conditioning.

\section{Experiments}
\label{sec:experiments}
In this section, we discuss the implementation details and the industrial dataset we used for our experiments. Finally, a thorough comparison between our approach and a state-of-the-art conditional diffusion model~\cite{layoutdiffusion} is assessed in terms of quality and consistency.

\begin{table*}[]
    \centering
    \resizebox{\linewidth}{!}{
    \begin{tabular}{r|cccc|cccc|ccc}
    & \multicolumn{4}{c}{\textbf{FID} $\downarrow$} & \multicolumn{4}{|c}{\textbf{KID} $\downarrow$} & \multicolumn{3}{|c}{\textbf{LPIPS} $\downarrow$} \\
    \textbf{Data} & $@2048$ & $@768$ & $@192$ & $@64$ & $@2048$ & $@768$ & $@192$ & $@64$ & \textit{AlexNet} & \textit{VGG-16} & \textit{SqueezeNet} \\ \midrule
    Synth \cite{layoutdiffusion} & $\mathbf{40.94}$ & $\mathbf{0.25}$ & $24.04$ & $6.77$ & $\mathbf{40.94}$ & $\mathbf{8.07}$ & $19 \times 10^3$ & $10 \times 10^3$ & $0.35$ & $0.49$ & $0.26$ \\
    Synth \textit{Ours} & $45.47$ & $0.30$ & $\mathbf{14.49}$ & $\mathbf{3.09}$ & $45.46$ & $8.73$ &  $\mathbf{10\times 10^3}$ & $\mathbf{3.6 \times 10^3}$ & $\mathbf{0.28}$ & $\mathbf{0.43}$ & $\mathbf{0.21}$ \\
    \end{tabular}
    }
    \caption{Assessment of generation quality. We report the FID and KID computed at different levels of the InceptionV3 \cite{inceptionv3} network, and the LPIPS computed with several backbones AlexNet \cite{alexnet}, SqueezeNet \cite{squeezenet} and VGG-16 \cite{vgg}.}
    \label{tab:fid-lpips}
\end{table*}

\subsection{Experimental setting}

\textbf{Diffusion model.} The proposed method follows the DDPM~\cite{ddpm} paradigm with a UNet~\cite{unet} architecture trained from scratch. We modified the input and output channels accordingly to support our bounding box encoding representation and the denoising of the segmentation map. Both during training and testing the number of denoising iterations were set to $1000$. We trained for $300$ epochs using AdamW~\cite{adamw} as optimizer with learning rate $1e^{-5}$ and batch size $8$ on two Nvidia RTX $4090$. The total training time is approximately $1$ day.

\vspace{0.3em}

\noindent \textbf{Downstream task.} For the semantic segmentation downstream task we emploied a UNet architecture with a ResNet-18~\cite{resnet} backbone. We used a single network for each segmentation class to avoid class balancing problems and concentrate on the synthetic data assessment. The training lasted $100$ epochs using AdamW as optimizer with learning rate $1e^{-5}$ and batch size $64$ on a single Nvidia RTX $4090$.

\vspace{0.3em}

\noindent \textbf{Dataset.} Since we focus on the industrial setting, we selected the Wood Defect Detection~\cite{openwood} dataset, a semantic segmentation and object detection collection of data for the wood manufacturing industry. It contains $20276$ images with semantic segmentation and bounding box annotations of $10$ different classes of wood defects. In our experiments, we decided to aggregate the $4$ classes of knots and avoiding the classes of blue stain and overgrown that are underrepresented. Thus, we obtained a dataset comprising of $20107$ images with a total of $5$ defect classes (knot, crack, quartzite, resin, marrow). 

Moreover, we split the dataset into three subsets: $70\%$ for training the diffusion model, $20\%$ for training the segmentation model, and $10\%$ as a fixed real test set. Additionally, the bounding box annotations from the $20\%$ real split are used to generate synthetic data for evaluating the semantic segmentation task. Figure~\ref{fig:dataset} illustrates some samples from the original dataset.

\begin{table}[]
    \centering
    \resizebox{\linewidth}{!}{
    \begin{tabular}{r|cccccc}
        & \multicolumn{6}{c}{\textbf{SAE ($\%$) $\downarrow$}} \\
        \textbf{Method} & Knot & Crack & Quartzite & Resin & Marrow & \textbf{Avg} \\
        \midrule
        Layout Diffusion \cite{layoutdiffusion} & $40.03$ & $83.82$ & $61.00$ & $85.59$ & $54.88$ & $46.77$ \\ 
        \textit{Ours} & $\mathbf{5.53}$ & $\mathbf{4.57}$ & $\mathbf{3.19}$ & $\mathbf{4.82}$ & $\mathbf{3.64}$ & $\mathbf{4.99}$ \\
    \end{tabular}
    }
    \caption{Comparison between our method and \cite{layoutdiffusion} in terms of Segmentation Alignment Error. The Avg is computed over all pixels.}
    \label{tab:sae}
\end{table}

\subsection{Data synthesis assessment}
To assess the quality of synthetic data, we compare our approach with the current state-of-the-art layout-conditional diffusion model \cite{layoutdiffusion}, utilizing its original code implementation and adapting it to take non-squared images. Specifically, we focus on evaluating the consistency between the generated defects and their corresponding bounding box constraints. To quantify this relationship, we introduce two metrics, the Segmentation Alignment Error (SAE) and the Empty Bounding-Box Rate (EBR).

\vspace{0.3em}

\textbf{Segmentation Alignment Error (SAE).}
With this measure we quantify how many generated defect pixels fall outside their designated bounding boxes, indicating misalignment between the generated defects and their constraints. Formally, let:
\begin{itemize}
    \item $\hat{P}$ be all the generated pixels of segmented defects,
    \item $\hat{P}_{out}$ be the generated pixels that fall outside the bounding boxes.
\end{itemize}
Thus, we define the metric as follows:
\begin{equation}
    \text{SAE} = \frac{\hat{P}_{out}}{\hat{P}}
\end{equation}
where a lower value indicates that the model is more consistent with the generation condition.

As shown in Table~\ref{tab:sae}, the method proposed in \cite{layoutdiffusion} struggles to maintain defect placement within the bounding boxes, resulting in a very high mean SAE of $46.77\%$ across all the defects. In contrast, our approach, leveraging a dual bounding box encoding strategy (BASD and C-BASD), significantly improves alignment, with only $4.99\%$ of generated pixels falling outside the given regions.

\vspace{0.3em}

\begin{table}[]
    \centering
    \resizebox{\linewidth}{!}{
    \begin{tabular}{r|cccccc}
        & \multicolumn{6}{c}{\textbf{EBR} ($\%$) $\downarrow$} \\
        \textbf{Method} & Knot & Crack & Quartzite & Resin & Marrow & \textbf{Avg} \\
        \midrule
        Layout Diffusion \cite{layoutdiffusion} & $13.43$ & $69.16$ & $48.76$ & $80.15$ & $28.44$ & $26.00$ \\ 
        \textit{Ours} & $\mathbf{0.86}$ & $\mathbf{2.41}$ & $\mathbf{4.98}$ & $\mathbf{2.22}$ & $\mathbf{0.89}$ & $\mathbf{5.51}$ \\
    \end{tabular}
    }
    \caption{Comparison between our method and \cite{layoutdiffusion} in terms of Empty Bounding-Box Rate. The Avg is computed over all bounding boxes.}
    \label{tab:ebr}
\end{table}

\textbf{Empty Bounding-Box Rate (EBR).}
To assess whether the generated defects correctly fall within their designated bounding boxes, we define the Empty Bounding-Box Rate (EBR). This metric quantifies how many bounding boxes remain empty, meaning no synthetic pixels are generated inside them. Formally, let:
\begin{itemize}
\item $B_{all} = \{b_k~|~b_k \in B_n~,~n=1,\dots,N\}$ be the set of all bounding boxes,
    \item $B_{miss} = \{b_k~|~b_k \in B_{all}~,~\mathcal{G} \cap b_k = \emptyset\}$ be the subset of bounding boxes that contain no generated pixels.
\end{itemize}
Thus, we define the metric as follows:
\begin{equation}
\text{EBR} = \frac{|B_{miss}|}{|B_{all}|}
\end{equation}
where higher values indicate that a larger number of bounding boxes have been missed during generation, signifying a poorer retrieval of the provided conditioning.

As reported in Table~\ref{tab:ebr}, the EBR metric shows the superiority of our proposal in retrieval abilities by a large margin. Specifically, our average EBR lies around 5.51\% on the total amount of bounding boxes and surpasses by more than 20\% points the competitor \cite{layoutdiffusion}. 

\vspace{0.3em}

\begin{figure*}
    \centering
    \includegraphics[width=0.98\linewidth]{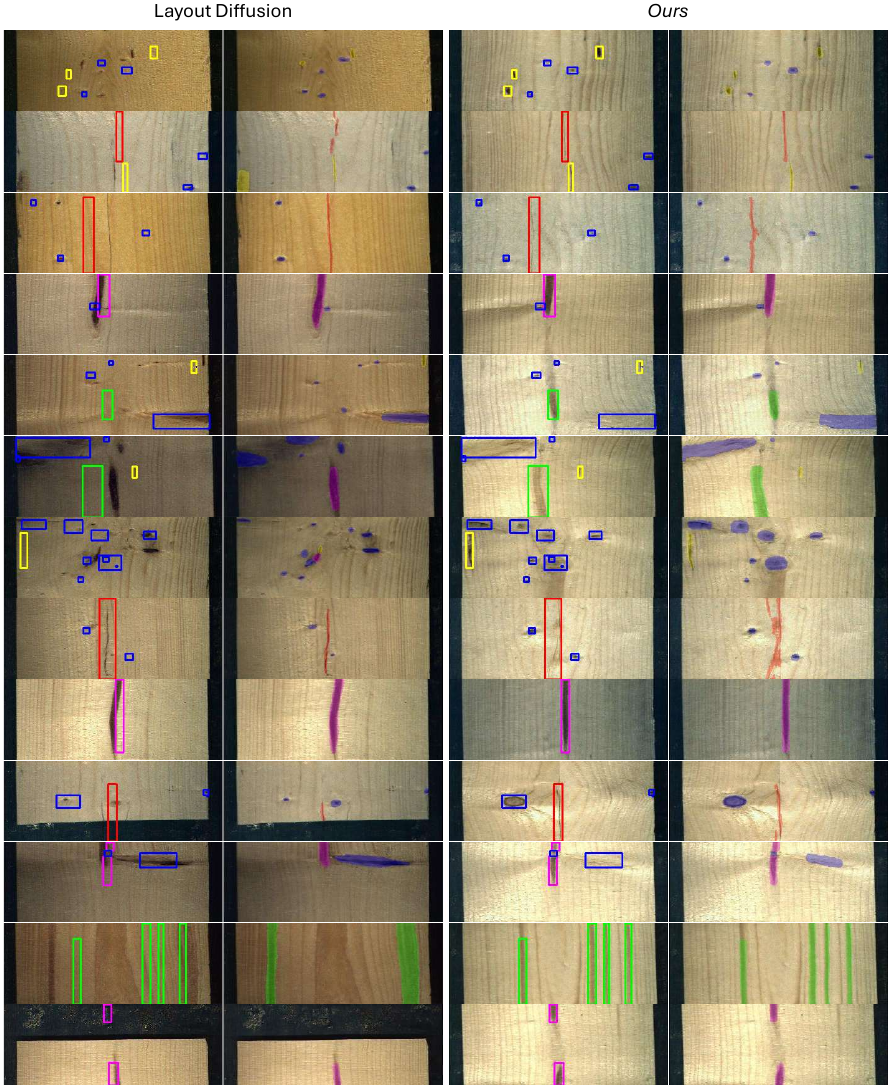}
    \caption{Qualitative comparison between our method and Layout Diffusion \cite{layoutdiffusion}. Each method shows the generated RGB image with respect to the bounding box condition - the same for both methods - and the overlapped defect segmentation map. The wood defects are the following: knot (blue), crack (red), quartzite (green), resin (yellow), marrow (magenta).}
    \label{fig:qualitative}
\end{figure*}

\textbf{Visual sample quality.} To further analyze the quality of the generated synthetic images, we report the Fréchet Inception Distance (FID) \cite{fid}, the Kernel Inception Distance (KID) \cite{kid} and the LPIPS \cite{lpips} metrics. As can be observed in Table~\ref{tab:fid-lpips}, our method tends to have better performance on low-level features with regard to the FID and KID metrics, meaning that the local perception of the details is better than the competitor. Additionally, our method outperforms \cite{layoutdiffusion} across all tested backbones in terms of LPIPS metric, confirming that the generated images exhibit higher perceptual realism across different network architectures.

\vspace{0.3em}

\textbf{Qualitative results.} To further illustrate this comparison, Figure~\ref{fig:qualitative} depicts qualitative examples. Moreover, the results demonstrate that \cite{layoutdiffusion} not only fails to confine defects within the bounding boxes but also occasionally generates wrong segmentation labels.

\subsection{Downstream task evaluation}
To evaluate the effectiveness of our synthetic data, we conduct a semantic segmentation experiment using a UNet architecture trained on different data configurations.

Starting from the 20\% split, we use the original bounding box annotations as guidance to generate couples of image and label. We do so for both methods, ours and \cite{layoutdiffusion}. We then use this synthetic split to train the segmentation pipeline. 
Moreover, to ensure a fair comparison between approaches, we discard synthetic pixel labels generated outside the bounding boxes conditioning. This prevents eventual generalization of the downstream segmentation given by extra synthetic labels generated without explicit conditioning.


\begin{table}[]
    \centering
    \resizebox{\linewidth}{!}{
    \begin{tabular}{r|cccccc}
        & \multicolumn{6}{c}{\textbf{F1} ($\%$) $\uparrow$} \\
        \textbf{Train data} & Knot & Crack & Quartzite & Resin & Marrow & \textbf{Avg} \\
        \midrule
        Real & $78.56$ & $48.80$ & $24.49$ & $45.00$ & $65.40$ & $52.45$ \\
        \midrule
        Synth \cite{layoutdiffusion} & $72.15$ & $8.20$ & $21.03$ & $18.01$ & $58.04$ & $35.49$ \\
        Synth \textit{Ours} & $76.57$ & $45.56$ & $12.82$ & $32.71$ & $58.18$ & $45.17$ \\
        \midrule
        Real+Synth \cite{layoutdiffusion} & $78.44$ & $46.71$ & $\mathbf{27.01}$ & $43.85$ & $\mathbf{71.09}$ & $53.42$ \\
        Real+Synth \textit{Ours} & $\mathbf{79.48}$ & $\mathbf{50.38}$ & $25.85$ & $\mathbf{46.29}$ & $66.11$ & $\mathbf{53.62}$ \\
    \end{tabular}
    }
    \caption{Downstream task assessment in terms of F1 score using real, synthetic and real+synthetic data during training.}
    \label{tab:f1}
\end{table}

Table~\ref{tab:f1} presents the F1 scores computed on the $10\%$ real test split, where we compare models trained on real data, synthetic data, and a combination of both. Notably, when training on synthetic data alone, our approach surpasses \cite{layoutdiffusion} by an impressive $10\%$, demonstrating its ability to generate more valid training samples. This highlights the superior quality and consistency of our synthetic segmentation maps, which provide a more reliable learning signal for the segmentation task.

When incorporating real data into the training process, the performance gap between the two methods narrows, as real samples provide a strong baseline. However, even in this hybrid setting, leveraging our synthetic data leads to the best overall F1 score, achieving a $+1.17\%$ improvement over using only real data. This result underscores the effectiveness of our method in complementing real-world annotations, reinforcing its practical utility in industrial applications where obtaining high-quality segmentation labels can be costly and time-consuming.
\section{Conclusion}
\label{sec:conclusion}

In this work, we are the first~\cite{review} to study the problem of data synthesis for semantic segmentation in industrial settings, where quality and precision is of importance. We devised a pipeline to generate synthetic RGB data and its segmentation label counterpart at the same time, starting from bounding box conditioning. This allows to decrease significantly the labeling costs while preserving the quality of the segmentation maps.

We validated the performances of our method by comparing our proposal with the current state-of-the art methodology adapted for the setting. We also assessed the quality of our generation through a downstream task, training a UNet with a combination of real and synthetic data.

The experiments suggest that our proposal is robust to spatial consistency generation, improving the performance of the downstream segmentation task. 

We also introduced dedicated metrics useful for the community to assess the correctness of layout-conditioned data generation.

{
    \small
    \bibliographystyle{ieeenat_fullname}
    \bibliography{main}
}


\end{document}